\documentclass[12pt]{article}
\pdfoutput=1
\usepackage{jcappub}
\usepackage{amsmath}
\usepackage{amssymb}
\usepackage{physics}
\usepackage{xcolor}
\usepackage{slashed}
\usepackage{subfigure}
\usepackage{float}
\usepackage{notoccite}

\definecolor{dgreen}{HTML}{008000}

\def\be{\begin{equation}}
\def\ee{\end{equation}}
\def\bea{\begin{eqnarray}}
\def\eea{\end{eqnarray}}

\title{Autonomous particles}

\author[1,2]{Nikola Andrejić}
\emailAdd{nikola.ing.nl@gmail.com}\author[1,3]{and Vitaly Vanchurin} 
\emailAdd{vitaly.vanchurin@gmail.com}
\affiliation[1]{Artificial Neural Computing, Weston, Florida, 33332, USA}
\affiliation[2]{University of Ni\v s, Faculty of Science and Mathematics, Vi\v segradska 33, Ni\v s, Serbia}
\affiliation[3]{Duluth Institute for Advanced Study, Duluth, Minnesota, 55804, USA}

\begin{document}

\abstract{
Consider a reinforcement learning problem where an agent has access to a very large amount of information about the environment, but it can only take very few actions to accomplish its task and to maximize its reward. Evidently, the main problem for the agent is to learn a map from a very high-dimensional space (which represents its environment) to a very low-dimensional space (which represents its actions). The high-to-low dimensional map implies that most of the information about the environment is irrelevant for the actions to be taken, and only a small fraction of information is relevant. In this paper we argue that the relevant information need not be learned by brute force (which is the standard approach), but can be identified from the intrinsic symmetries of the system. We analyze in details a reinforcement learning problem of autonomous driving, where the corresponding symmetry is the Galilean symmetry, and argue that the learning task can be accomplished with very few relevant parameters, or, more precisely, invariants. For a numerical demonstration, we show that the autonomous vehicles (which we call autonomous particles since they describe very primitive vehicles) need only four relevant invariants to learn how to drive very well without colliding with other particles. The simple model can be easily generalized to include different types of particles (e.g. for cars, for pedestrians, for buildings, for road signs, etc.) with different types of relevant invariants describing interactions between them. We also argue that there must exist a field theory description of the learning system where autonomous particles would be described by fermionic degrees of freedom and interactions mediated by the relevant invariants would be described by bosonic degrees of freedom. This suggests that the effectiveness of field theory descriptions of physical systems might be connected to the learning dynamics of some kinds of autonomous particles, supporting the claim that the entire universe is a neural network.}

\maketitle
\section{Introduction} \label{sec:intro}

Broadly speaking there are three interrelated problems of autonomous driving \cite{DNA22}. The first problem is environment detection which is done through complex systems including cameras and different types of sensors \cite{JYVHJBJW21}. The second problem is decision-making which is based on the input from the environment \cite{LLYL21,DTLZ22,SF22}. The third problem is producing the desired real-world behaviour of the autonomous vehicle which is the subject of vehicle control \cite{KLL22}. In this paper we assume that the first and third problems are solved (although we briefly discuss the third problem at the end of Sec. \ref{sec:Learning}) and focus on solving the second problem, i.e. decision-making. The decision-making problem is an example of a reinforcement learning task where an agent, an autonomous vehicle, has to take an action (e.g. accelerate in some direction) based on the information about its environment (e.g. positions and velocities of all other vehicles and of its destination). Then the main decision-making task for the autonomous vehicle is to  model a map from a very high-dimensional space of the environment to a very low-dimensional space of possible actions. Evidently, the decision-making must involve coarse-graining of information about the environment by keeping only a small amount of relevant information. The standard (or brute force) approach is to use a neural network to learn the relevant information. In this paper we argue that the relevant information must be described by invariants (or scalars) with respect to Galilean (and permutation) transformations. This allows us to identify a small number of relevant invariants (only four invariants) that are sufficient for the autonomous vehicle to learn how to drive very well without colliding with other vehicles. 

To better understand the proposal, consider a test charged particle moving in an electromagnetic field created by other charged particles (sources). If we know the position and velocity of the test particle at some time $t$, then to determine its position and velocity at $t+dt$ all we need to know is the electric, $\vec{E}$, and magnetic, $\vec{B}$, fields at time $t$ at the position of the test particle. This means that not all information (i.e. fine-grained description of the dynamical state) of the sources is required for the test particle to ``decide'' how to move while obeying the laws of electrodynamics. The parameters, $\vec{E}$ and $\vec{B}$, already  contain the relevant coarse-grained description of all other particles which is both necessary and sufficient for the test particle to move according to Lorentz force law regardless of how many source particles there are. 

The above example can be understood in at least two ways: local and global. The first is the traditional (or local) field-theoretic point of view, where sources emit local bosons according to some law and the only relevant information that affects the test particle is the strength of the bosonic (e.g. electromagnetic) field and all of the fined-grained information about the dynamical state of the sources is completely irrelevant. In fact, all of the experimentally verified laws of motion (e.g. Lorentz force law) are produced by interactions between fermionic particles (e.g. electrons) mediated by local bosonic fields (e.g. electromagnetic field). However, we can also consider the second point of view, where the test particle is an agent (or autonomous particle) which is making decisions of how to move based on a global information about its environment. In the global view, the test particle first scans all of its environment to determine a fined-grained (possibly up to some precision)  dynamical state of all other particles and then decides how to coarse grain this information and how to move accordingly. In electrodynamics, for an electron, this would be equivalent to determining the position and velocity of every other charged particle in the universe, then deciding to calculate local $\vec{E}$ and $\vec{B}$ fields according to Maxwell's equations, and, finally, deciding what should its acceleration be according to these calculations and its own dynamical state (i.e. calculating the Lorentz force). 

It should be evident, that the two pictures, i.e. local and global, are equivalent (or dual) at least for the case of electrodynamics. While the local picture is a lot more useful for developing physical theories where the locality and symmetries play the key role, the global picture seems to be more useful for solving reinforcement learning tasks such as autonomous driving where the identification of the relevant information (relevant for driving) is of central importance. To make the analogy more apparent, one should think of the autonomous particle as an autonomous vehicle or a car, which must decide both what information is relevant for achieving its goals as well as how to move according to this information. In other words it can decide what kind of ``bosons'' it sees and what kind of ``fermion'' it considers itself to be. However, the main difference is that for the particle physics problem the dynamics of both bosons and fermions is determined by some local Lagrangian and for the autonomous driving problem the dynamics must be determined by some loss function and therefore should be learned. In this respect one can consider the problem of ``autonomous particles'' as more general than ``fundamental particles'' in a sense that their dynamics need not be fixed and can be learnt.

On a more practical level, the field-theoretic picture may be used to  tackle, for example, the autonomous driving problem in the following way. For simplicity, let us assume that each autonomous particle can, in principle, know the position and velocity of all particles (i.e. other cars or obstacles) in its environment. Then its main task is to use this information and, possibly, its destination, to calculate a local coarse-grained information which consists of only a few parameters whose total number is very small and fixed regardless of how many cars are in the environment. It is important to emphasise that the relevant parameters are both untrained and invariant under symmetry transformations (e.g. rotations, shifts, permutations of other particles), but they are used as input to a neural network whose output is the desired position (or, equivalently, velocity or acceleration) in the next time step. In addition, it is also assumed that the loss function is defined as a function of the invariants only. As we will see, only four relevant invariants and only thirty neurons is enough to train an unconstrained neural network \cite{VV2021} how to drive fairly well without colliding with other particles. 

The paper is organized as follows.  In Sec. \ref{Sec:Problem} first formulate the problem of decision-making in context of autonomous driving and then describe a neural network architecture which can be used to obtain a solution. In Sec. \ref{sec:Learning} we discuss general aspect  of the learning dynamics such as identification of relevant invariants and construction of the loss function. In Sec. \ref{Sec:Numerics} we describe in details the numerical simulation and discuss numerical results. In Sec. \ref{Sec:Discussion} we summarize main results and discuss implications of the results for machine learning and fundamental physics.

\section{Problem and solution}\label{Sec:Problem}

Consider a collection of ``autonomous particles'' (cars, aircrafts, robots, etc.) with positions and velocities $\vec{r}_\alpha(t)$ and $\vec{v}_\alpha(t)$ ($\alpha = 1,2,\ldots, {\cal N}$) moving in an arbitrary number of dimensions $d$. Each particle has its own main objective of reaching its destination $\vec{R}_\alpha$ as fast as possible while avoiding collisions with other autonomous particles. At each time step each particle needs to calculate its own position at time step $t+\varepsilon$ (or velocity, or acceleration) such that these objectives are met based on the information that is available at time $t$. The objective for different particles may be conflicting with each other and each particle should be making decentralized decisions of where to accelerate and thus how to move. Furthermore, we assume that each autonomous particle knows, with arbitrary precision, relative positions and velocities of all other particles at some time $t$ and position of its destination. Using this information (and, if applicable, information about its own velocity with respect to preferred coordinate system, e.g. Earth) the particle $\alpha$ constructs a small number of Galilean (e.g. rotational and transnational) and permutation (of all other particles) invariants $\phi_{\alpha i}$'s where $i=0,\ldots n-1$. Based on values of $\phi_{\alpha i}$'s the particle decides where to move on the next time step (e.g. what should be its velocity). As we will see in the following sections, these decisions are made by independent neural networks (one  for each particle) that take $\phi_{\alpha i}$'s as inputs, and these networks are trained by minimizing their respective loss functions, i.e. $H_\alpha(\phi_{\alpha 0},  \phi_{\alpha 1}, ... , \phi_{\alpha n-1})$.

\begin{figure}[H]
\centering
\includegraphics[scale=0.44]{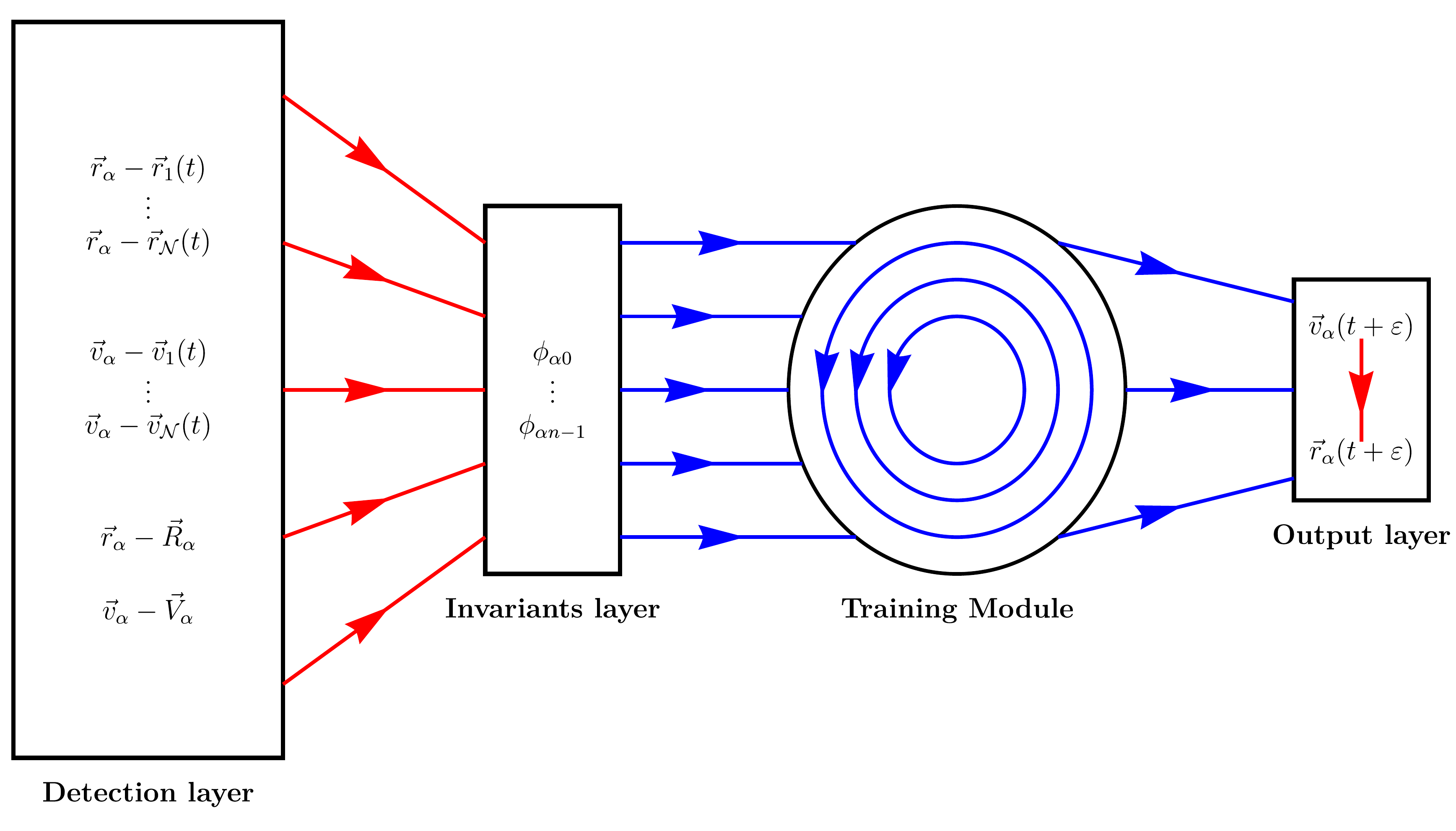}
\caption{Neural network architecture with red arrows indicate non-trainable connections while the blue arrows indicate trainable connections.}\label{fig:nnarch}
\end{figure}

We propose a particular architecture of a neural network for each particle that can be divided into four units (see Fig.~\ref{fig:nnarch}). The first unit is what we call a detection layer. The detection layer contains information (about all other particles and of the destination) which is passed in a feed forward way to invariants layer. The invariants layer has one neuron for each invariant $\phi_{\alpha i}$ and its state is equal to the value of the corresponding invariant. Connections between the detection layer and invariants layer are non trainable and  determined by the functional form of the invariants $\phi_{\alpha i}$. The feed-forward structure (detection layer) $\to$ (invariants layer) is responsible for the coarse-graining of data by filtering irrelevant information and keeping only invariants that would be relevant for making decisions. The invariants, $\phi_{\alpha i}$'s, can be considered as an input layer for the trainable part of the network or what we call the training module. The final unit is the output layer that tells the ``autonomous particle'' where to move (e.g. what the particle's position, velocity or acceleration should be). The feed forward connections between the invarinats layer and the training module, between the training module and the output layer and all of the connections within the training module are trainable variables.

More precisely, the detection layer for particle $\alpha$ can takes as inputs all of relative positions and velocities, i.e. 
\be
\left (\vec{r}_\alpha(t) - \vec{r}_1(t),\vec{v}_\alpha(t) - \vec{v}_1(t), ..., \vec{r}_\alpha(t) - \vec{r}_{\cal N}(t),\vec{v}_\alpha(t) - \vec{v}_{\cal N}(t),  \vec{r}_\alpha(t) - \vec{R}_\alpha,  \vec{v}_\alpha(t) - \vec{V}_\alpha\right ),
\ee
where for simplicity we shall assume that all destinations are at rest, i.e. $\vec{V}_\alpha=0$. Then all connections between  the detection layer and the invariants layer are assumed to be such that appropriate invariants are calculated and identified with states of the corresponding neurons,   
\be\label{eq:invariants}
\phi_{\alpha i}(t) = \sum_{\beta\neq\alpha}  \psi_i(\vec{r}_\alpha(t) - \vec{r}_\beta(t),\vec{v}_\alpha(t) - \vec{v}_\beta(t), \vec{r}_\alpha(t) - \vec{R}_\alpha,  \vec{v}_\alpha(t)-\vec{V}_\alpha).
\ee
All $\psi_i$'s and $\phi_{\alpha i}$'s are scalars (i.e. invariant) with respect to Galilean (e.g. rotation, shifts) transformations and in addition $\phi_{\alpha i}$'s are invariant under permutations of all other particles, i.e. $\beta\neq\alpha$. In Sec. \ref{Sec:Numerics} we construct a concrete example of the invariants where $\psi_i$'s are functions of only two vectors $\vec{r}_\alpha(t) - \vec{r}_\beta(t)$ and $\vec{v}_\alpha(t) - \vec{v}_\beta(t)$.

From the invariants layer, the states of neurons are fed to the training module and then from the training module to  $\vec{r}_\alpha(t+\varepsilon)$ and/or $\vec{v}_\alpha(t+\varepsilon)$ in the output layer. Assuming constant acceleration during the interval $[t,t+\varepsilon]$ and in order to be kinematically consistent, we impose
\begin{equation}\label{ralphat+1}
    \vec{r}_\alpha(t+\varepsilon) = \vec{r}_\alpha(t) + \frac{\varepsilon}{2}\left(\vec{v}_\alpha(t) + \vec{v}_\alpha(t+\varepsilon)\right),
\end{equation}
i.e. the architecture of the output layer is such that we first calculate $\vec{v}_\alpha(t+\varepsilon)$ based on the input layer through trainable connections and only then we calculate $\vec{r}_\alpha(t+\varepsilon)$ through non trainable connections described in Eq.~\ref{ralphat+1}. Moreover, given that particle's input velocity is $\vec{v}_\alpha(t)$, to produce desired outputs $\vec{r}_\alpha(t+\varepsilon)$ and $\vec{v}_\alpha(t+\varepsilon)$ it must accelerate at
\begin{equation}\label{eq:acceleration}
\vec{a}_\alpha(t)=\frac{\vec{v}_\alpha(t+\varepsilon)-\vec{v}_\alpha(t)}{\varepsilon}.
\end{equation}
If we think of \eqref{eq:acceleration} as a part of some final activation function, then we may say that, in effect, we train for the constant acceleration (or force) of the particle during the time interval $[t,t+\varepsilon]$ as it will uniquely determine $\vec{v}_\alpha(t+\varepsilon)$ and $\vec{r}_\alpha(t+\varepsilon)$.

\section{Learning dynamics}\label{sec:Learning}

During learning the trainable variables of the training module are adjusted such that the outputs $\vec{r}_\alpha(t+\varepsilon)$ and $\vec{v}_\alpha(t+\varepsilon)$ meet the learning objectives. The simplest objectives that we can impose are that the cars avoid collisions and that they arrive at their destination in the smallest number of time steps. To achieve these goals, we consider a loss function $H_\alpha$ which depends on a small number of global invariants \eqref{eq:invariants}  and a single local Galilean invariant,
\be
\phi_{\alpha 0}\left ( \vec{r}_\alpha(t) - \vec{R}_\alpha , \vec{v}_\alpha(t) - \vec{V}_\alpha, \vec{r}_\alpha(t+\varepsilon) - \vec{R}_\alpha, \vec{v}_\alpha(t+\varepsilon) - \vec{V}_\alpha \right),
\ee
which is built out of only local (to the particle in question)  vector quantities.  

As already mentioned, the loss function must be expressible as 
\begin{equation}
    H_\alpha = H_\alpha(\phi_{\alpha0},\phi_{\alpha 0}, ..., \phi_{\alpha n-1}),
\end{equation}
where $\phi_{\alpha i}$ are introduced in Eq.~\eqref{eq:invariants}. In other words, other particle's positions and velocities can appear in the loss $H_\alpha$ only in a coarse-grained way through global invariants $\phi_{\alpha i}$'s. The main difference is that, unlike evaluating invariants for the activation dynamics, for the learning dynamics (if possible) all $\phi_{\alpha i}$'s should be evaluated at time step $t+\varepsilon$ in order for the particle $\alpha$ to make decisions such that its loss at $t+\varepsilon$ is minimal. However, due to decentralized learning, the neural network of particle $\alpha$ cannot influence nor has access to the decisions that the neural networks of other particles $\beta\neq\alpha$ would make. We address this problem by calculating estimates of the positions and velocities denoted by
\bea
\vec{\rho}_\beta &\approx& \vec{r}_\beta(t+\varepsilon)\\
\vec{u}_\beta &\approx& \vec{v}_\beta(t+\varepsilon) 
\eea
These estimates can be made at different orders with zeroth order given by
\begin{equation}\label{eq:estimate0}
\vec{u}_\beta = \begin{cases}
\vec{v}_\alpha(t + \varepsilon) & \beta=\alpha\\
\vec{v}_\beta(t) & \beta\neq\alpha
\end{cases}\qquad \vec{\rho}_\beta =\begin{cases}
\vec{r}_\alpha(t + \varepsilon) & \beta=\alpha\\
\vec{r}_\beta(t) & \beta\neq\alpha.
\end{cases}
\end{equation}
and first order given by
\begin{equation}\label{eq:estimate}
\vec{u}_\beta = \begin{cases}
\vec{v}_\alpha(t+\varepsilon) & \beta=\alpha\\
\vec{v}_\beta(t) & \beta\neq\alpha
\end{cases}\qquad \vec{\rho}_\beta =\begin{cases}
\vec{r}_\alpha(t+\varepsilon) & \beta=\alpha\\
\vec{r}_\beta(t) + \varepsilon\vec{v}_\beta(t) & \beta\neq\alpha.
\end{cases}
\end{equation}
which is the order that we shall use for producing numerical results in Sec. \ref{Sec:Numerics}. Another interesting possibility is to train a neural network (either separately or as an additional layer that would be added between detection and invariants layers, see Fig. \ref{fig:nnarch}) for each particle whose main task would be to better predict the positions and velocities of other particles and thus improve the estimates further. Moreover, if all neural networks of the particles are interconnected (which is not the case in the problem that we consider), then there could be a centralized decision making and the overall loss function can depend on the data at $t+\varepsilon$ for all other particles and we can use $\vec{r}_\beta(t+\varepsilon)$ and $\vec{v}_\beta(t+\varepsilon)$ for these estimates. 

Learning dynamics of trainable variables of each neural network proceeds via the gradient descent method,
\begin{equation}\label{backprop}
d\vec{Q}_\alpha = -\eta_\alpha\frac{\partial H_\alpha}{\partial \vec{Q}_\alpha},
\end{equation}
where $\vec{Q}_\alpha=(w_{\alpha ij},b_{\alpha i})$ is a collective notation for of all trainable variables (weights and biases) in the particle $\alpha$ neural network and $\eta_\alpha$ is its learning rate. By employing Eq.~\eqref{ralphat+1} we can rewrite the gradient by explicitly extracting the outer-most factor from back-propagation expansion
\begin{equation}\label{eq:backpropstepone}
    \frac{\partial H_\alpha}{\partial Q_{\alpha i}}= -\eta\left(\frac{1}{2}\frac{\partial H_\alpha}{\partial\vec{\rho}_\alpha} + \frac{\partial H_\alpha}{\partial\vec{u}_\alpha}\right)\cdot\frac{\partial\vec{u}_\alpha}{\partial Q_{\alpha i}}.
\end{equation}
This is where we first encounter the gradients of the loss function with respect $\vec{\rho}_\alpha$ and $\vec{u}_\alpha$. Since these are the only two vectors that are calculated through activation dynamics in the neural network, only they are considered as dynamical variables from the standpoint of the back-propagation while all other quantities appearing in $H_\alpha$ are held fixed. To calculate these gradients we may (and in our numerical example will) need to reuse data from the detection layer.

Let's explore what kind of loss function $H_\alpha$ can be in order to satisfy our learning objective. If the main learning objective, of a given autonomous particle $\alpha$, is to arrive at its destination as fast as possible while avoiding collisions with other particles, then the loss function $H_\alpha$ must contain two kinds of terms:
\begin{itemize}
\item terms which describe interaction of a particle with its destination and
\item terms which describe interaction of a particle with all other particles.
\end{itemize}

The first kind of terms must include at least one long range invariants just because interactions with the destination must be significant at arbitrary distances or, otherwise, destination may never be reached. (Note, that terms of the first kind may also be short range if, for example, the learning objective is not only to go to a destination, but also to slow down once the particle is close to the destination.) Some interesting examples for the first kind of terms are 
\begin{equation}
\phi_{\alpha 0} = \abs{\vec{\rho}_\alpha-\vec{R}_\alpha}\label{eq:term1}
\end{equation}
or
\begin{equation}
\phi_{\alpha 0} = \vec{u}_\alpha\cdot\frac{\vec{\rho}_\alpha-\vec{R}_\alpha}{\abs{\vec{\rho}_\alpha-\vec{R}_\alpha}}.\label{eq:term2}
\end{equation}
Minimization of either one of these terms brings the particle closer to its destination. Gradients of \eqref{eq:term1} with respect $\vec{\rho}_\alpha$ and of \eqref{eq:term2} with respect to $\vec{u}_\alpha$ are the same and equal to the unit vector along $\vec{\rho}_\alpha-\vec{R}_\alpha(t)$, but \eqref{eq:term2}  has the advantage of being bounded and this is the term we use in our simple numerical model (see Sec. \ref{Sec:Numerics}).

The second kind of terms, that we also call collision-preventing terms, are intrinsically non-local since they would involve a sum over all particles \eqref{eq:invariants}. However, if we only want local configurations of particles to influence significantly the behaviour of our test particle then certain constraints must apply to its asymptotic  behavior of $\psi_i$'s. For example, if particles are distributed uniformly in a $d$-dimensional space, then the total contribution of all collision-preventing terms must be convergent which means that the gradients of $\psi_{i}$'s in Eq.~\eqref{eq:invariants} with respect to both $\vec{v}_\alpha-\vec{v}_\beta$ and $\vec{r}_\alpha-\vec{r}_\beta$ must decay as \footnote{Note that from the standpoint of back-propagation 
$$\frac{\partial}{\partial(\vec{r}_\alpha-\vec{r}_\beta)}=\frac{\partial}{\partial\vec{r}_\alpha}$$
and similar for the velocities.} 
\bea
\frac{\partial \psi_{i}}{\partial (\vec{r}_\alpha-\vec{r}_\beta)} &\lesssim& |\vec{r}_\alpha-\vec{r}_\beta|^{-d} \\
\frac{\partial \psi_{i}}{\partial(\vec{v}_\alpha-\vec{v}_\beta)} &\lesssim& |\vec{r}_\alpha-\vec{r}_\beta|^{-d}.
\label{eq:convergence} 
\eea 
This guarantees that the gradient terms $\frac{\partial H_\alpha}{\partial\vec{Q}_\alpha}$ (i.e. their contribution from particle-particle interaction), that are responsible for learning dynamics of $\vec{Q}_\alpha$ (see Eqs. \eqref{backprop} and \eqref{eq:backpropstepone}), are not dominated by far away particles, i.e. our test particle $\alpha$ does not need to know much about their positions and velocities. Moreover, to prevent collisions, terms of the second kind must be increasing and, possibly, diverging when two particles come really close to each other. However, the exact form of the collision-preventing terms depends on the specifics of the ``avoiding collisions'' part of the learning task, such as the shape and size of particle, the locality of the behaviour (i.e. how far away other objects have to be so that the moving particle starts making relevant decision to avoid them), etc.

Before we proceed to the numerical part of the paper, let's us briefly emphasise that in any real world applications the autonomous vehicles cannot control directly the output velocity and position. More precisely, what the computer of an autonomous vehicles can control are some parameters like the position of the steering wheel, the gear, how to press the gas pedal etc. These parameters do determine kinematic quantities and how to predict the real-world reaction (behaviour) of the system to change in these parameters falls within the scope of vehicle control. An interesting, physics-informed take on this problem can be found in Ref.~\cite{KLL22}. Within our framework this problem can be integrated by splitting the output layer into layer of controlled parameters and a ``technological'' layer that would reproduce correctly the behaviour of the vehicle. This layer can be another, possibly physics informed, unit of the neural network that is either pre-trained or trained in real time together with our training module. In the following section we describe a simple numerical model where a primitive form of the ``technological'' layer is completely contained within the neural network architecture that calculates $\vec{u}_\alpha$. 

\section{Numerical simulation}\label{Sec:Numerics}

In this section we describe a numerical simulation of $\cal N$ autonomous vehicles, or cars, on a roadless 2D driving polygon. This simulation is a special case of the model of autonomous particles described in the previous sections as well as of a more general autonomous particles model described in Sec. \ref{Sec:Discussion}. 

In the initial conditions, position $\vec{r}_\alpha(0)$ and destination $\vec{R}_\alpha$ for all cars are initialized randomly on a square polygon, and all initial velocities are set to zero, $\vec{v}_\alpha(0) =0$. Orientation of each cars is randomly initialized and during motion it is always assumed to point along its velocity $\vec{v}_\alpha(t)$. When a car arrives at its destination, $\vec{R}_\alpha$ is reinitialized to another random point on the same 2D driving polygon. The size of the time step is taken to be $\varepsilon=1$ and we enforce that the maximum allowed speed of any car is set equal to $1$, see Eq.~\eqref{eq:velocityact}.

We use a neural network architecture described in Sec. \ref{Sec:Problem} and Fig.~\ref{fig:nnarch} which is the same for all cars. Trajectory of each car is constructed in the following way:\begin{enumerate}
    \item detection layer of each car detects the positions and velocities of all cars at time $t$ and from these positions and its own destination, the car constructs the relevant invariants $\phi_{\alpha i}$ for the input layer;
    \item the invariants are fed to the training module and produce the output; 
    \item this first forward pass of the data and the first output created is only used to perform back-propagation i.e. to update weights and biases; 
    \item after the update, we perform another forward pass of the data but now with updated parameters and the outputs that we get are interpreted to be the position and the velocity of the particle at time $t+\varepsilon$;
\end{enumerate}
Then we repeat all steps from (1) to (4) with $t$ incremented by $\varepsilon$. In principle, each car can move (i.e. self-drive) for as many time steps as desired by repeating above loop and by randomly generating a new destination whenever an old destination is reached. 

Since we only use four invariants, the invariants layer will have only four neurons. The output layer by default has also four neurons for four components of $\vec{r}_\alpha(t+\varepsilon)$ and $\vec{v}_\alpha(t+\varepsilon)$. The training module has 24 additional neurons, themselves fully connected in an unconstrained way as described in \cite{VV2021}. This means that we have $(24+4)(24+2) = 728$ trainable weights and $24+2 = 26$ trainable biases or $754$ trainable variables in total. The activation function for the neurons in the training module is taken to be Rectified Linear Unit (ReLU), while the activation on the output neurons that calculate $\vec{v}_\alpha(t+\varepsilon)$ is designed in a certain way to ensure that there is a maximum allowed speed. Namely, if  $w_1$ and $w_2$ are the pre-activation inputs of the two neurons that calculate two components of $\vec{v}_\alpha(t+\varepsilon)$, then we take
\begin{equation}\label{eq:velocityact}
    \vec{v}_\alpha(t+\varepsilon) = \mathcal{R}\frac{\sigma(|\vec{z}|)}{|\vec{z}|}\vec{z},\quad \vec{z} = \left(\tanh\frac{w_1}{5},\tanh\frac{w_2}{5}\right),
\end{equation}
where $\sigma$ denotes the usual sigmoid activation function and $\mathcal{R}$ rotates the velocity from the frame defined by car's orientation (where the local $x$ axis is along $\vec{v}_\alpha(t)$) to a global frame (or the rest frame of destinations) that is fixed at all times. In this way the speed is effectively connected to a sigmoid activation, while the ratio of the two components of $\vec{z}$ determines the change in the direction away from $\vec{v}_\alpha(t)$. Note that with this activation function, pre-activation input $\vec{w}$ uniquely determines the velocity at $t+\varepsilon$ in the rotated frame with $x$-axis along $\vec{v}_\alpha(t)$. We choose to use this kind of activation function to enforce maximum speed as well as to ensure that our primitive car-like object have a preferred orientation, i.e. there is a front and there a back of a car. With this we break a symmetry in a sense that to accelerate in the direction of motion as well as perpendicular to the direction of motion, themselves being different, separately remain the same no matter what the orientation of the car is with respect to the stationary frame.\footnote{In principle, there could be autonomous objects that are equally easy/hard to accelerate in all directions no matter what their current velocity is and then their output activation functions do not need to encode information about orientation.} However, this activation function doesn't allow our primitive car-like objects to have reverse nor it is able to replicate many other features of a real car such as the minimal radius of curvature of the trajectory, gears, brakes and so on. All of these problems fall within the scope of vehicle control or, in our terms, ``technological'' layer that was mentioned in the previous section. 

\begin{figure}[H]
    \centering
    \subfigure[]{\fbox{\includegraphics[width=0.25\textwidth]{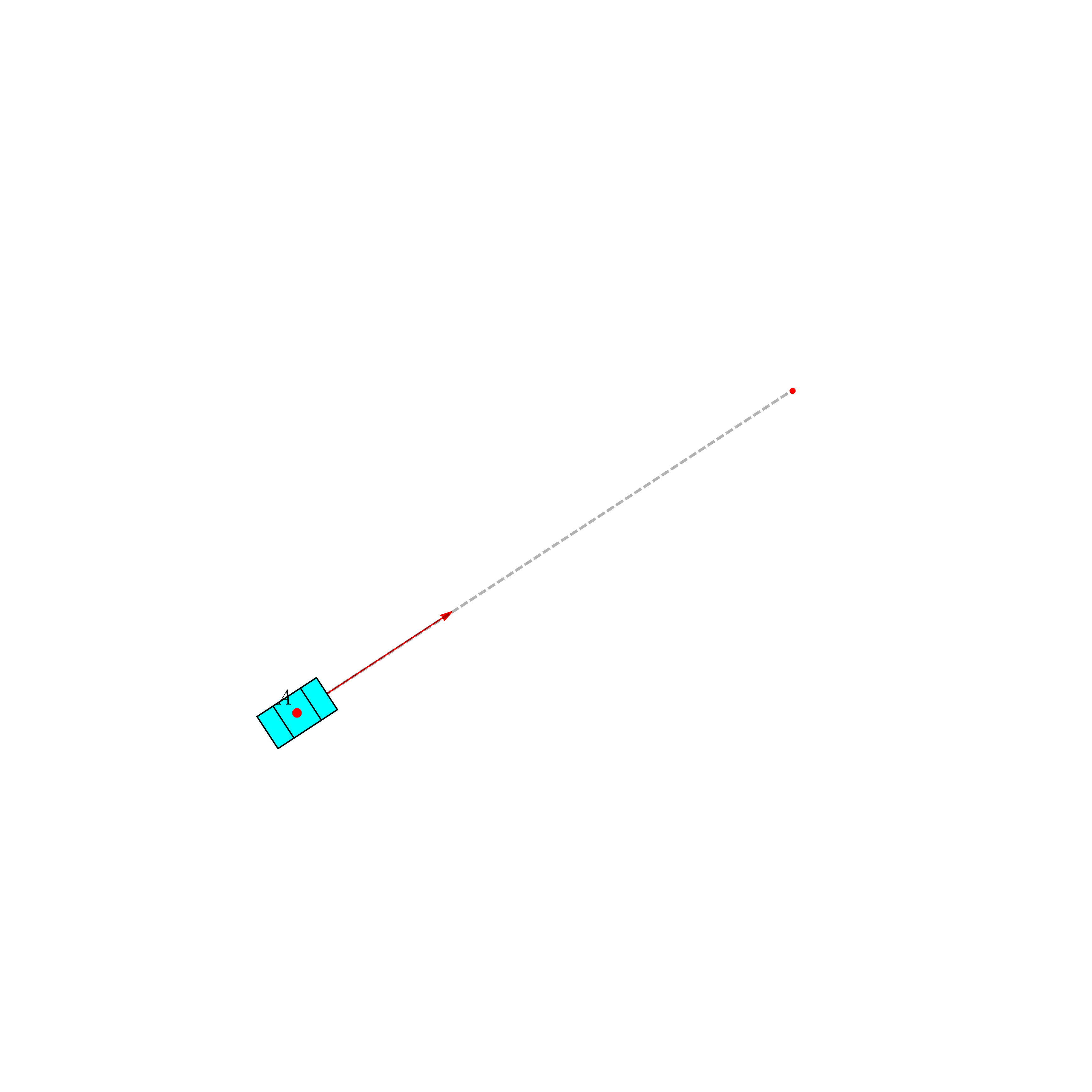}}}
    \subfigure[]{\fbox{\includegraphics[width=0.25\textwidth]{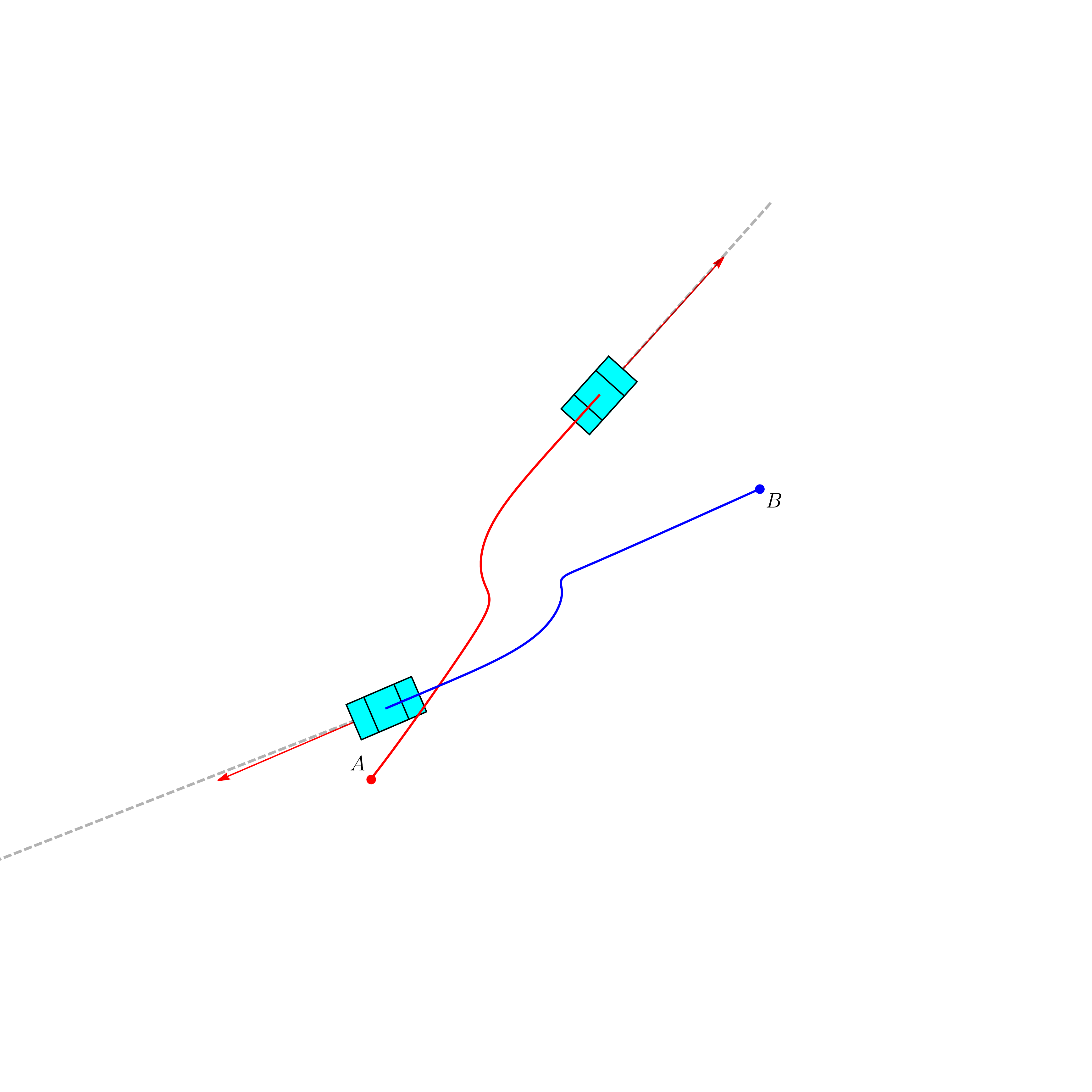}}} \\
    \subfigure[]{\fbox{\includegraphics[width=0.25\textwidth]{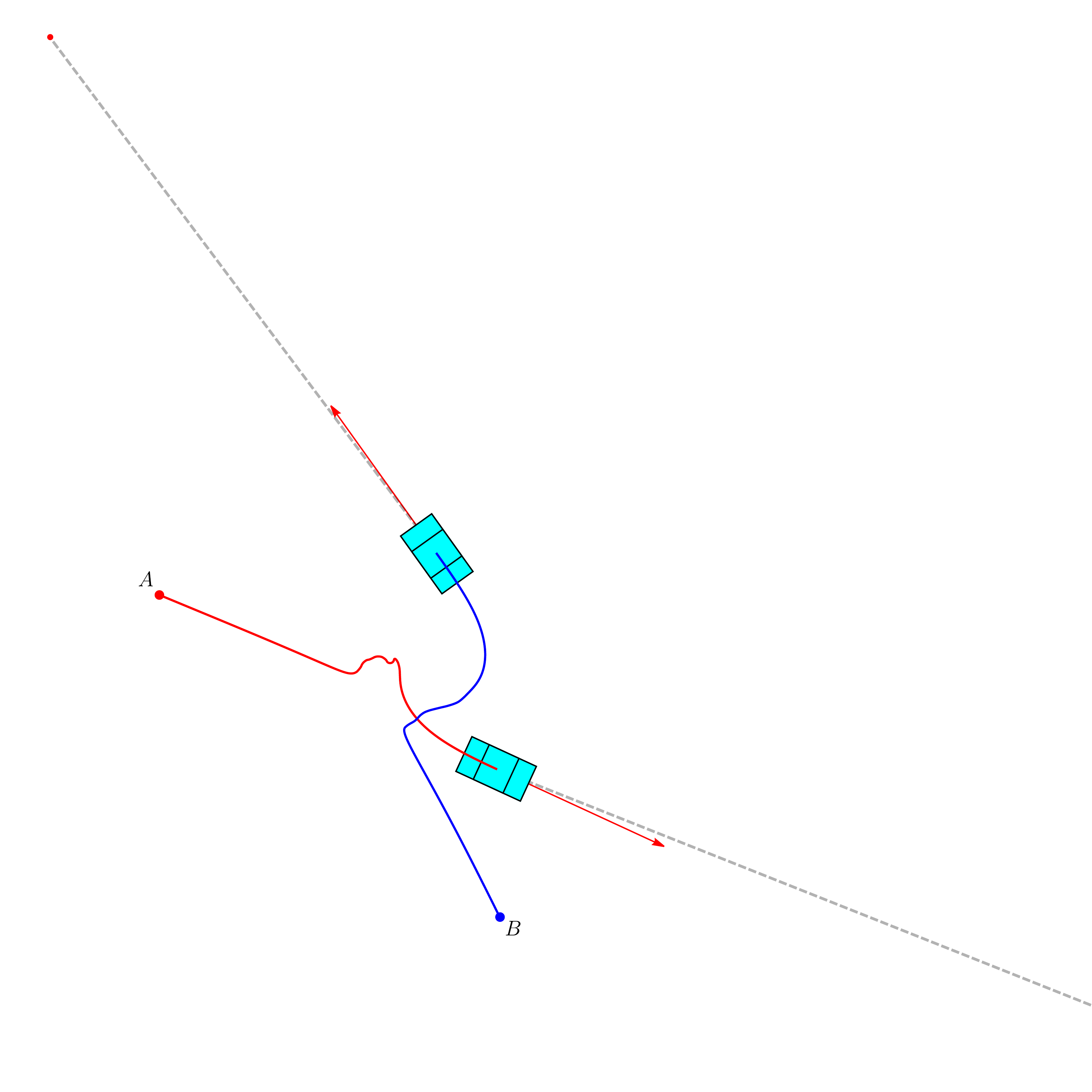}}}
    \subfigure[]{\fbox{\includegraphics[width=0.25\textwidth]{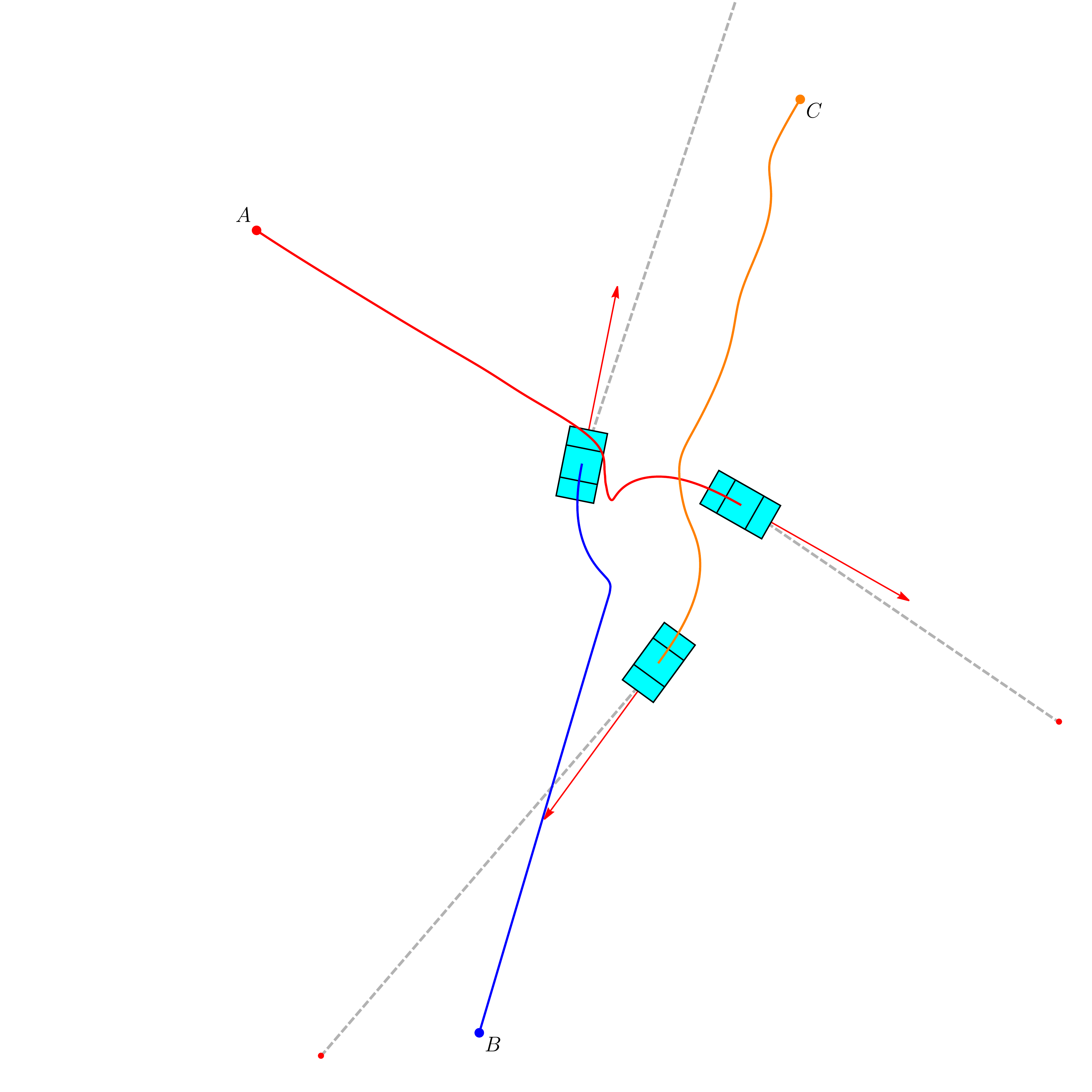}}}
    \caption{(a) When far away from other cars, the car's velocity (red arrow) is perfectly aligned with the direction to its destination (dashed line); (b) Two cars meet and both of them turn left in their own frame to avoid the collision. Their respective trajectories are shown as well as their positions some time after the encounter; (c) Two cars meet but now one of them turns left (red trajectory) and the other one turns right (blue trajectory). Situation is resolved when one of them slows down (red trajectory) and the other one persists (blue trajectory). The one that slows down goes behind the one that persists. (d) Three cars meet and go around each other until the traffic congestion is resolved.}
    \label{fig:resol}
\end{figure}

The main goal for each car is to arrive at the destination while not colliding with other cars and this goal is encoded in the loss function $H_\alpha$ of the neural network of car $\alpha$, that is, achieving the goal should correspond to minimizing $H_\alpha$. For the particular example that we present the loss function has the following form:
\begin{equation}\label{eq:loss}
H_\alpha = \phi_{\alpha 0} + \phi_{\alpha 1} + \phi_{\alpha 2} 
\end{equation}
where, 
\bea
\phi_{\alpha0} &=& (\vec{u}_\alpha - \vec{V}_\alpha) \cdot\frac{\vec{\rho}_\alpha-\vec{R}_\alpha}{\abs{\vec{\rho}_\alpha-\vec{R}_\alpha}}+ {\left(1-\left | \vec{u}_\alpha - \vec{V}_\alpha \right |^2\right)^{- \frac{1}{2}}}, \label{eq:phi0}\\
\phi_{\alpha1} &=& \sum_{\beta\neq\alpha} \psi_1(\vec{\rho}_\alpha(t)-\vec{\rho}_\beta(t), \vec{u}_\alpha(t)-\vec{u}_\beta(t)) = \lambda_1 \sum_{\beta\neq\alpha}\frac{1}{\abs{\vec{\rho}_\alpha -\vec{\rho}_\beta}^3},\label{eq:phi1}\\
\phi_{\alpha2} &=& \sum_{\beta\neq\alpha} \psi_2 (\vec{\rho}_\alpha(t)-\vec{\rho}_\beta(t), \vec{u}_\alpha(t)-\vec{u}_\beta(t)) = - \lambda_2 \sum_{\beta\neq\alpha}\frac{(\vec{u}_\alpha-\vec{u}_\beta)\cdot(\vec{\rho}_\alpha-\vec{\rho}_\beta)}{\abs{\vec{\rho}_\alpha -\vec{\rho}_\beta}^4}.\label{eq:phi2}
\eea
The hyperparameters $\lambda_1$ and $\lambda_2$ determine the relative importance of each term in Eq.~\eqref{eq:loss} and $(\vec{u}_\beta,\vec{\rho}_\beta)$ is given by Eq.~\eqref{eq:estimate}. Motivations behind each term in the loss function can be summarized as follows: 
\begin{itemize}
    \item $\phi_{\alpha 0}$ describes the long range interaction between the car and its destination and, when minimized, makes the velocity of the car to point in the direction of the destination and to be as large as possible. We rely on this term to make sure that the car eventually arrive to its destination.
\item $\phi_{\alpha 1}$ prevents the cars from coming too close to each other, it quickly blows up whenever any car approaches car $\alpha$ while its minimum is when car $\alpha$ is far away, at infinity, from all other cars.
\item $\phi_{\alpha 2}$ prevents the cars from both coming close to each other and moving head on when they are close. 
\end{itemize}
Both $\phi_{\alpha 1}$ and $\phi_{\alpha 2}$ describe short range car-car interactions and have the purpose of preventing collisions. It is easy to see that the decay of their gradients satisfies the locality property \eqref{eq:convergence}. In $d=2$ dimensions $\psi_{i}$'s must decay faster than $|\vec{r}_\alpha-\vec{r}_\beta|^{-1}$ and in Eqs. \eqref{eq:phi1} and \eqref{eq:phi2} both $\psi_{1}$ and $\psi_{2}$ decay as $|\vec{r}_\alpha-\vec{r}_\beta|^{-2}$.
Therefore we expect that, when car $\alpha$ is far away from other cars, $\phi_{\alpha 0}$ dominates the loss and its priority is to align its velocity with the direction pointing towards its destination. On the other hand, when there are other cars nearby the two other terms $\phi_{\alpha 1}$ and $\phi_{\alpha 2}$ would attempt to prevent collisions. Also note the second term in $\phi_{\alpha 0}$ (similar to the $\gamma$-factor from special relativity), although not essential for the learning tasks has proven to be very useful for stabilizing the behaviour of cars in our simulation. It prevented cars from moving at near maximum speed at all times by changing the relative preference of tangential versus centripetal acceleration in favour of the tangential component. This preference is highly skewed towards the centripetal acceleration when the car moves at near maximum speed because there we enter the zone of vanishing gradient of the sigmoid.
The expected behaviour of the collection of cars is numerically verified in the simulation that we run for this example (see Fig.~\ref{fig:resol}).

For the invariants layer (see Fig. \ref{fig:nnarch}) we use two invarinats which appear directly in the loss function, i.e. $\phi_{\alpha 1}$ and $\phi_{\alpha 2}$, and then two more invariant that can be extracted from $\phi_{\alpha 0}$, i.e. 
\be
    \phi_{\alpha3}= \vec{v}_\alpha(t)\cdot\frac{(\vec{r}_\alpha(t)-\vec{R}_\alpha)}{\abs{\vec{r}_\alpha(t)-\vec{R}_\alpha}}
\ee
and
\be
     \phi_{\alpha4} = \left | \vec{v}_\alpha(t) -\vec{V}_\alpha \right | = {v}_\alpha(t).
\ee
This is a minimal set of $\phi_{\alpha i}$'s such that the loss function can be expressed as a function of them. However, nothing stops us from taking additional invariants for the invariants layer which may be useful for learning efficiency, i.e. the change in their values might influence the loss function significantly indicating their relevance for the learning task (see Sec. \ref{Sec:Discussion}).

Numerical simulation was performed for ${\cal N}=50$ autonomous particles, or cars, with loss functions given by Eq. \eqref{eq:loss} and hyperparameters set to $\lambda_1=\lambda_2=10^6$. The magnitudes of $\lambda_1$ and $\lambda_2$ in effect control the effective size of each car, i.e. the characteristic distance between the cars within which the contribution to the total gradient $\partial H_\alpha/\partial \vec{Q}_\alpha$ of the collision preventing terms $\phi_{\alpha 1}$ and $\phi_{\alpha 2}$ is comparable with the contribution of the other term $\phi_{\alpha 0}$, i.e. the one that makes the car go to the destination. We made an arbitrary choice to represent our cars as identical rectangles of size $260\times140$ and have chosen $\lambda_{1,2}$ accordingly by trial and error to enforce that the effective size corresponds to the visual size of the car. The learning rates are fixed at $\eta_\alpha = 0.5$.

In the simulation, each car always arrives at its destination and, when far away from other cars, it moves directly towards it. When it meets another car it will try to steer to the side to avoid it while slowing down slightly. Depending on its first experience of avoiding the collision, it will learn to either steer to the left or to the right learning left-hand or right-hand traffic rule. Then in any subsequent close encounter with another car it will try to steer to that side to avoid collision. The other car involved might learn the same rule or the opposite one and the two situations will be handled differently with the resolution of the encounter being much more efficient if they learned the same rule. Another interesting situation that arises when three or more cars need to cross their roads and these situations are resolved by roundabout-like behavior. In Fig.~\ref{fig:resol} we show the four selected examples described above. A full animation of the simulation is available at \cite{animation}.

\begin{figure}[H]
    \centering
    \subfigure[]{\includegraphics[width=0.45\textwidth]{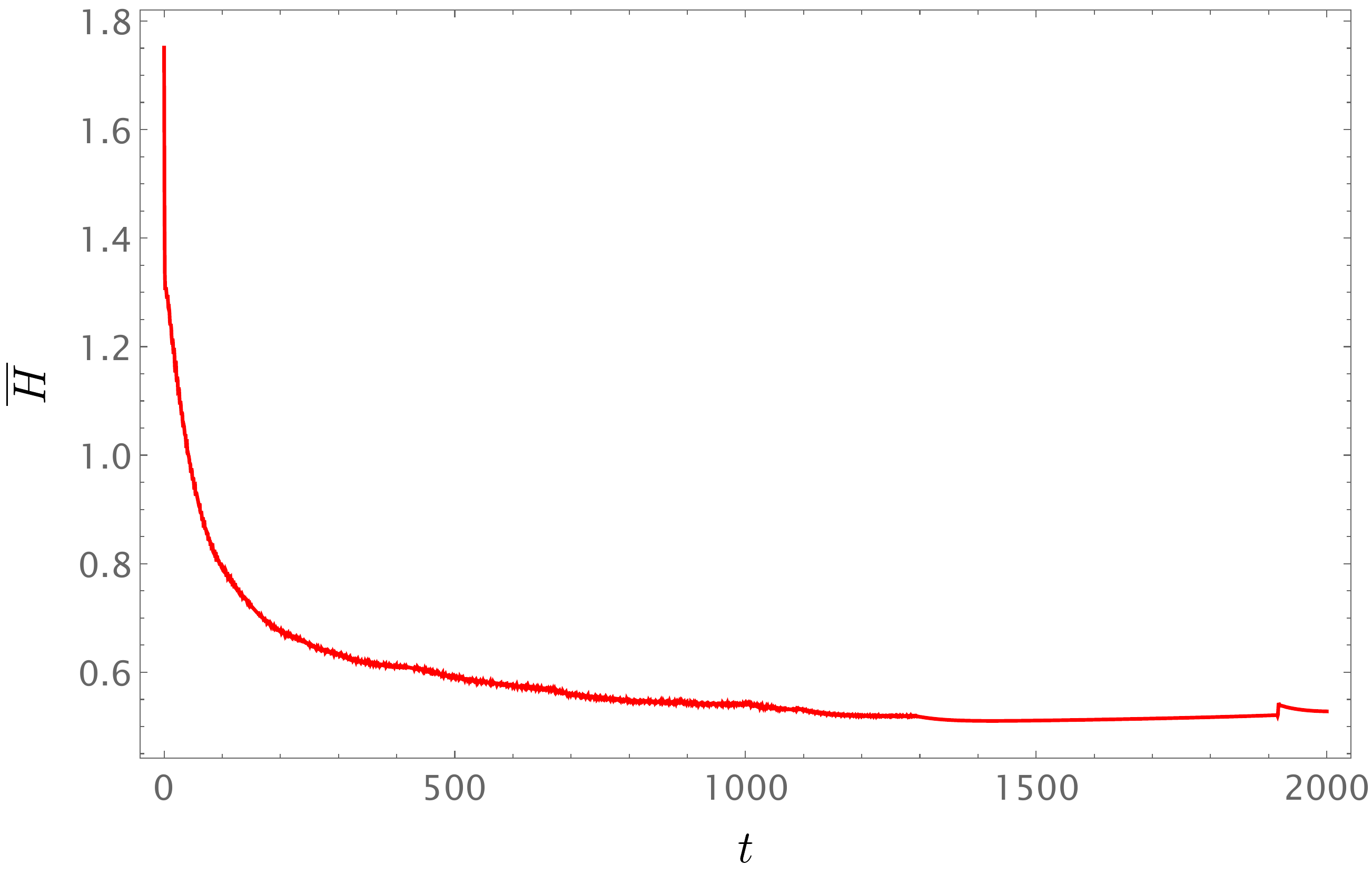}}
    \subfigure[]{\includegraphics[width=0.45\textwidth]{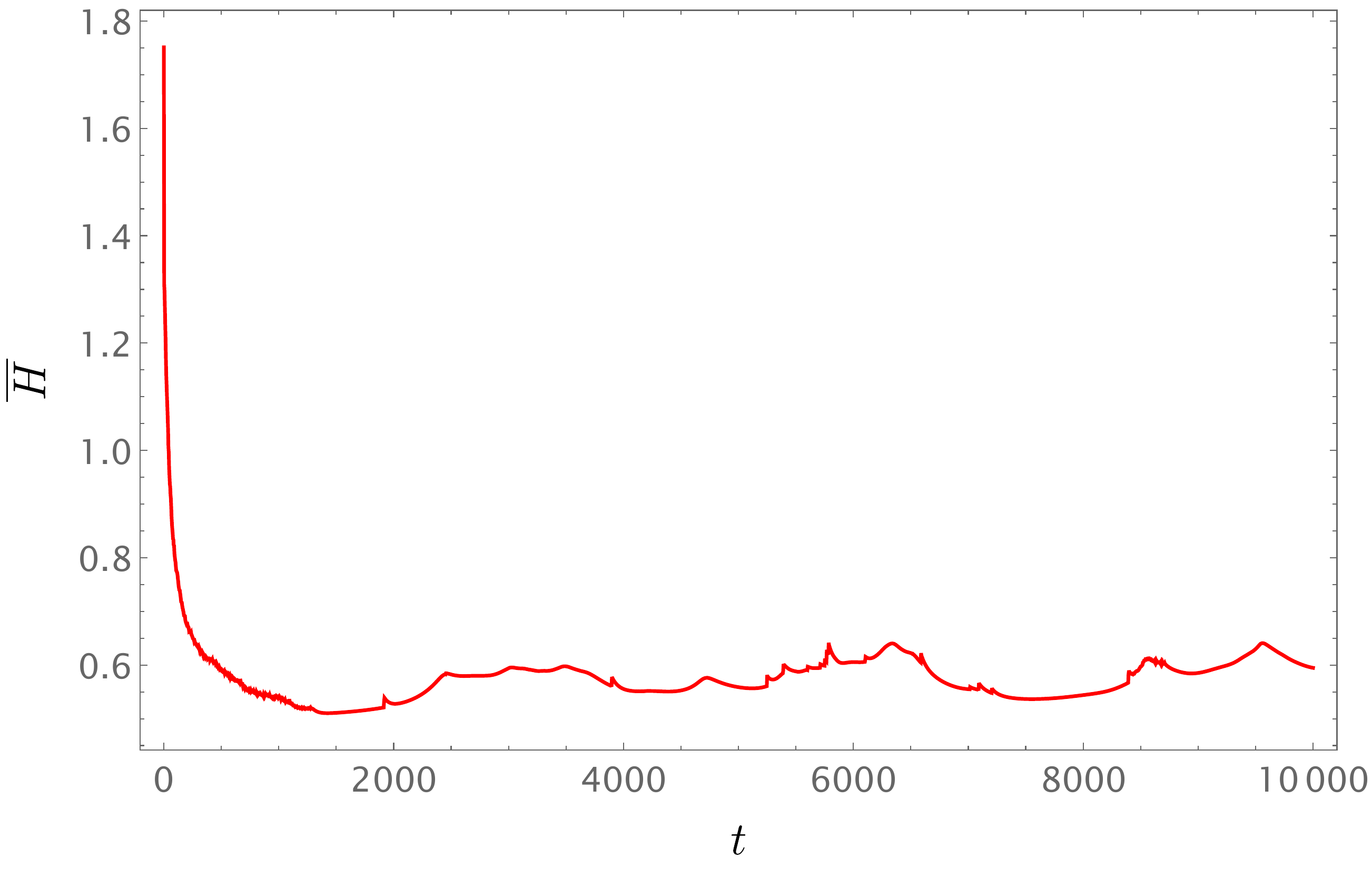}} \\
    \subfigure[]{\includegraphics[width=0.55\textwidth]{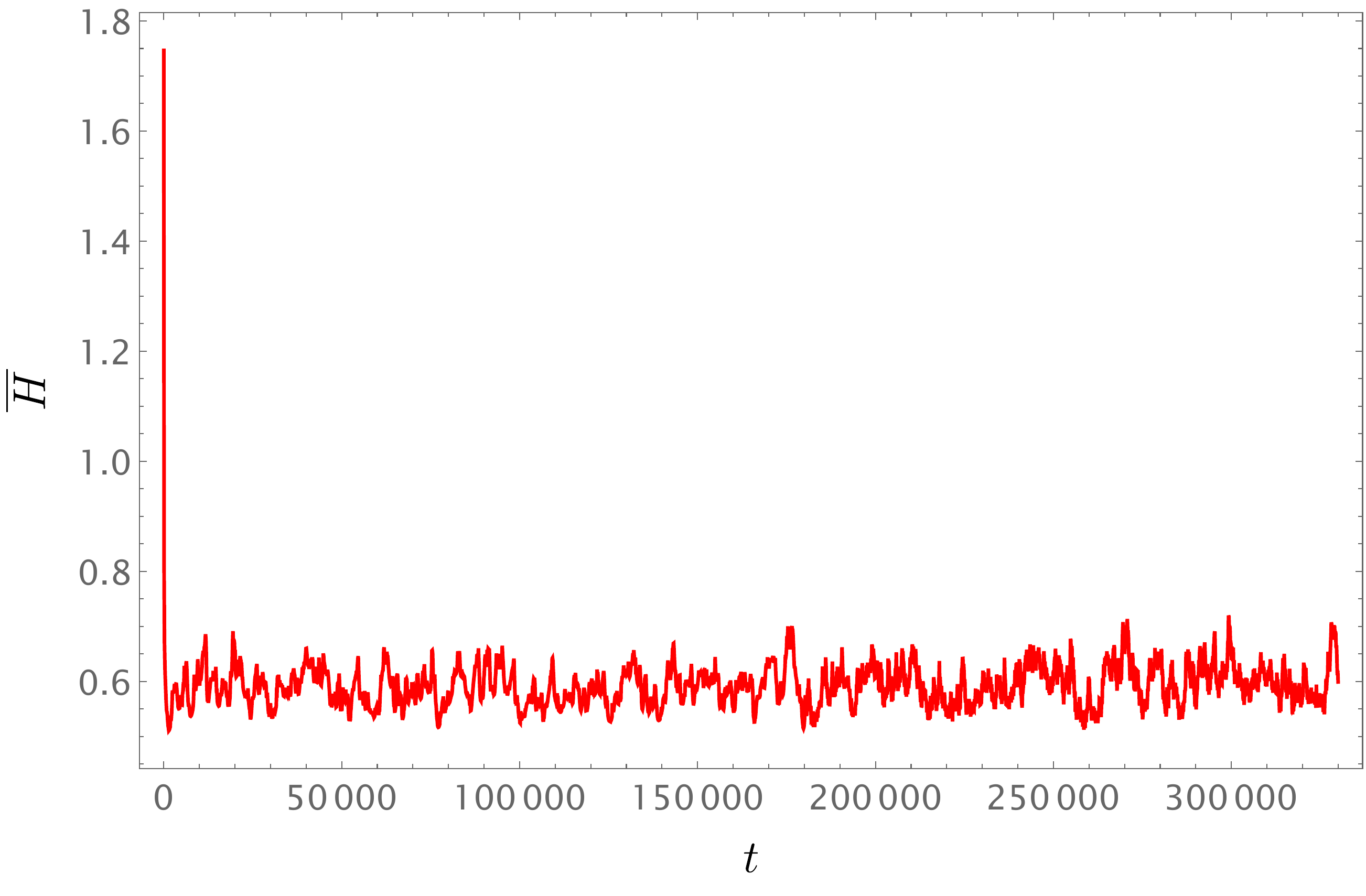}}
    \caption{(a) The initial stage of the exponential decay of the average loss with characteristic time numerically determined to be $\tau\approx 100$. (b) Magnification of the first few fluctuation cycles. (c) Evolution of the average loss for the entire duration of the simulation.}
    \label{fig:lossevolv}
\end{figure}

By following the time evolution of the loss function averaged over all cars, 
\be
\overline{H} =\frac{1}{\cal N} \sum_{\alpha=1}^{\mathcal{N}}H_\alpha,
\ee
one can clearly see that the average loss goes down very quickly. In Fig.~\ref{fig:lossevolv} we show the dependence of the average loss on time with different magnifications of the time axis. After that, the average loss fluctuates with no apparent change in the amplitude. Evidently, the cars learn basic behavior for achieving their main goals (i.e. arriving to destination and avoiding collisions) very fast, over the first $~1000$ time steps, and if necessary cars can also quickly relearn their behaviors (e.g. switching from right-hand to left-hand traffic rule or vise versa). The ability to quickly learn and relearn is the main advantage of the proposed architecture and is due entirely to a small number of relevant invariants (only four) and a small number of neurons (only thirty) used by each autonomous car. In a more realistic simulation, we expect that the total number of relevant invariants could be larger, but the main idea of training using invariants should remain the same.

\section{Discussion}\label{Sec:Discussion}

In this article, we proposed a solution to the problem of decision-making in reinforcement learning tasks such as autonomous driving. We have shown, by performing analytical and numerical calculations, that the decision-making should be based on only relevant information that does not need to be learned by brute force, but that can be identified from the intrinsic symmetries of the system, e.g. Galilean or permutation symmetry. Although the original motivation was coming from the use of symmetries in constructing physical theories, such as electrodynamics, the symmetry consideration proved to be extremely useful for a simplified problem of autonomous driving. This raises a question of how the framework can be generalized to include more realistic autonomous vehicles such as self-driving cars, air-crafts, robots, etc. 

There are few directions along which the model of autonomous particles can be generalized and very likely improved. First of all, the number of relevant invariants, which was only four for the numerical simulation (see Sec. \ref{Sec:Numerics}), can be increased to include other and more general terms in the invariants layer (see Sec. \ref{Sec:Problem}) as well as in the loss function (see Sec. \ref{sec:Learning}). Secondly, the number of different types of particles can be arbitrary (e.g. for cars, for pedestrians, for buildings, for road signs, etc.) and interactions between different types can be described by different relevant invariants. And finally, for identification of the relevant invariants (which is the key step in our construction) it may be useful to develop an effective filed theory description of the (different types of) autonomous particles where the relevant invariants would be nothing but Green's functions for respective particles/sources. 

Indeed, the invariants that were used for both activation and learning dynamics \eqref{eq:phi1} and \eqref{eq:phi2} can be interpreted as a sum over $\delta$-function like densities at the location of other autonomous particles weighted by some invariant functions. It is not immediately clear if these invariant functions would be Green's functions for some fields, nor that the invariants that were used are the most relevant invariants for the problem at hand. However, this highlights an interesting field-theoretic aspect of the autonomous driving problem that may turn out to be useful. In particular, it may be interesting to see if there exist an effective field theory description with fermions describing autonomous particles and bosons describing interactions between them, so that the bosonic Green's function describes the relevant invarinats and the corresponding force on fermions is the force that would be learned by the autonomous particles. In this respect, even in our simple self-driving problem, a particular field theory would emerges where the cars learn what kind of fermions they have to be in order to better meet the learning objective.

This raises another question, namely, can we formulate the learning task for the autonomous particle such that the known (or unknown) field theories would emerge? For example, can we formulate a learning task for an autonomous particle such that electrodynamics emerges as an outcome of the learning dynamics? And if the answer is affirmative and some known field theories can indeed be described as a learning dynamics, then this would give additional support to the claim that the entire universe is a neural network \cite{VV2020NN}. Of course, this would not explain why the laws of physics on microscopic scales are governed by quantum mechanics  (see, however, \cite{VV2020NN, katsnelson, vanchurin2}) and on astrophysical scales by the general theory of relativity (see, however, \cite{VV2020NN, vanchurin2}), but it would allow us to develop an alternative (or dual) picture of physical systems in the context of the learning theory. The standard ``physical'' picture would be to write down a Lagrangian function which describes fermions interacting with each other by means of bosons, and the dual ``learning'' picture would be to write down a loss function which describes learning objective for autonomous particles (or fermions) interacting with each other by means of relevant invariants (or bosonic Green's functions). We leave all these and other related questions for future research.

\bibliographystyle{unsrt}
\bibliography{references}

\end{document}